\RequirePackage{xcolor}
\documentclass[9pt,shortpaper,twoside,web]{ieeecolor}
\usepackage{generic}
\usepackage{cite}
\usepackage{amsmath,amssymb,amsfonts}
\usepackage{algorithmic}
\usepackage{graphicx}
\usepackage{textcomp}
\usepackage{booktabs}
\usepackage{array}
\usepackage{stfloats}
\usepackage{url}
\usepackage{verbatim}
\usepackage{hyperref}
\usepackage{bm}
\usepackage{makecell}
\usepackage[nohyperlinks,nolist]{acronym}
\usepackage{colortbl}
\graphicspath{{figures/}}  

\def\BibTeX{{\rm B\kern-.05em{\sc i\kern-.025em b}\kern-.08em
    T\kern-.1667em\lower.7ex\hbox{E}\kern-.125emX}}
\usepackage{etoolbox}
\makeatletter
\@ifundefined{color@begingroup}%
  {\let\color@begingroup\relax
   \let\color@endgroup\relax}{}%
\def\fix@ieeecolor@hbox#1{%
  \hbox{\color@begingroup#1\color@endgroup}}
\patchcmd\@makecaption{\hbox}{\fix@ieeecolor@hbox}{}{\FAILED}
\patchcmd\@makecaption{\hbox}{\fix@ieeecolor@hbox}{}{\FAILED}

\begin{document}
\title{Robust Path Planning via Learning from Demonstrations for Robotic Catheters in Deformable Environments}
\author{Zhen Li, Chiara Lambranzi, Di Wu, Alice Segato, Federico De Marco, Emmanuel Vander Poorten, \\ Jenny Dankelman, and Elena De Momi, \IEEEmembership{Senior Member, IEEE}
\thanks{This work was supported by the ATLAS project. This project has received funding from the European Union’s Horizon 2020 research and innovation programme under the Marie Sklodowska-Curie grant agreement No 813782.}
\thanks{Zhen Li is with the Department of Electronics, Information and Bioengineering, Politecnico di Milano, 20133 Milan, Italy, and also with the Department of Biomechanical Engineering, Delft University of Technology, 2628 CD Delft, The Netherlands (e-mail: zhen.li@polimi.it).}
\thanks{Chiara Lambranzi is with the Department of Electronics, Information and Bioengineering, Politecnico di Milano, 20133 Milan, Italy, and also with Istituto Italiano di Tecnologia, 16163 Genova, Italy (e-mail: chiara.lambranzi@polimi.it).}
\thanks{Di Wu is with the Department of Mechanical Engineering, Katholieke Universiteit Leuven, 3000 Leuven, Belgium, and also with the Faculty of Mechanical, Maritime and Materials Engineering, Delft University of Technology, 2628 CD Delft, The Netherlands (e-mail: di.wu@kuleuven.be).}
\thanks{Alice Segato was with the Department of Electronics, Information and Bioengineering, Politecnico di Milano, 20133 Milan, Italy (e-mail: alice.segato@polimi.it).}
\thanks{Federico De Marco is with Centro Cardiologico Monzino, IRCCS, 20138 Milan, Italy (e-mail: federico.demarco@gmail.com).} 
\thanks{Emmanuel Vander Poorten is with the Department of Mechanical Engineering, Katholieke Universiteit Leuven, 3000 Leuven, Belgium  (e-mail: emmanuel.vanderpoorten@kuleuven.be).}
\thanks{Jenny Dankelman is with the Department of Biomechanical Engineering, Delft University of Technology, 2628 CD Delft, The Netherlands (e-mail: j.dankelman@tudelft.nl).}
\thanks{Elena De Momi is with the Department of Electronics, Information and Bioengineering, Politecnico di Milano, 20133 Milan, Italy (e-mail: elena.demomi@polimi.it).}
}

\maketitle

\begin{abstract}
\textit{Objective:} Navigation through tortuous and deformable vessels using catheters with limited steering capability underscores the need for reliable path planning. State-of-the-art path planners do not fully account for the deformable nature of the environment. \textit{Methods:} This work proposes a robust path planner via a learning from demonstrations method, named Curriculum Generative Adversarial Imitation Learning (C-GAIL). This path planning framework takes into account the interaction between steerable catheters and vessel walls and the deformable property of vessels. \textit{Results:} \textit{In-silico} comparative experiments show that the proposed network achieves a 38\% higher success rate in static environments and 17\% higher in dynamic environments compared to a state-of-the-art approach based on GAIL. \textit{In-vitro} validation experiments indicate that the path generated by the proposed C-GAIL path planner achieves a targeting error of 1.26$\pm$0.55mm and a tracking error of 5.18$\pm$3.48mm. These results represent improvements of 41\% and 40\% over the conventional centerline-following technique for targeting error and tracking error, respectively. \textit{Conclusion:} The proposed C-GAIL path planner outperforms the state-of-the-art GAIL approach. The \textit{in-vitro} validation experiments demonstrate that the path generated by the proposed C-GAIL path planner aligns better with the actual steering capability of the pneumatic artificial muscle-driven catheter utilized in this study. Therefore, the proposed approach can provide enhanced support to the user in navigating the catheter towards the target with greater accuracy, effectively meeting clinical accuracy requirements. \textit{Significance:} The proposed path planning framework exhibits superior performance in managing uncertainty associated with vessel deformation, thereby resulting in lower tracking errors. 
\end{abstract}


\begin{acronym}[Orocos]
    \acro{pci}[PCI]{Percutaneous Coronary Intervention}
    \acro{dfs}[DFS]{Depth First Search}
    \acro{birrt}[bi-RRT]{bi-directional Rapidly-exploring Random Tree}
    \acro{aco}[ACO]{Ant Colony Optimization}
    \acro{ga}[GA]{Genetic Algorithm}
    \acro{lfd}[LfD]{Learning from Demonstrations}
    \acro{drl}[DRL]{Deep Reinforcement Learning}
    \acro{rl}[RL]{Reinforcement Learning}
    \acro{or}[OR]{Operating Room}
    \acro{drunet}[DRU-Net]{Deep Residual U-Net}
    \acro{cta}[CTA]{Computed Tomography Angiography}
    \acro{ccm}[CCM]{Centro Cardiologico Monzino}
    \acro{mdct}[MDCT]{Multidetector Computed Tomography}
    \acro{ecg}[ECG]{Electrocardiogram}
    \acro{em}[EM]{Electromagnetic}
    \acro{fov}[FoV]{Field-of-View}
    \acro{dsc}[DSC]{Dice Similarity Coefficient}
    \acro{ddf}[DDF]{Dense Displacement Field}
    \acro{roi}[ROI]{Region of Interest}
    \acro{RRT}[RRT]{Rapidly-exploring Random Tree}
    \acro{stn}[STN]{Spatial Transform Network}
    \acro{cnn}[CNN]{Convolutional Neural Network}
    \acro{pbd}[PBD]{Position-based Dynamics}
    \acro{pam}[PAMs]{Pneumatic Artificial Muscles}
    \acro{ftl}[FTL]{Follow-The-Leader}
    \acro{bc}[BC]{Behavioral Cloning}
    \acro{gail}[GAIL]{Generative Adversarial Imitation Learning}
    \acro{cgail}[C-GAIL]{Curriculum Generative Adversarial Imitation Learning}
    \acro{GAN}[GAN]{Generative Adversarial Network}
    \acro{gui}[GUI]{Graphical User Interface}
    \acro{bc}[BC]{Behavioral Cloning}
    \acro{ppo}[PPO]{Proximal Policy Optimization}
    \acro{ivus}[IVUS]{IntraVascular UltraSound}
    \acro{oct}[OCT]{Optical Coherence Tomography}
    \acro{fbg}[FBG]{Fiber Bragg Grating}
    \acro{cto}[CTO]{Chronic Total Occlusion}
    \acro{cl}[CL]{Curriculum Learning}
    \acro{dof}[DoF]{Degree of Freedom}
    \acro{pso}[PSO]{Particle Swarm Optimization}
    \acro{dmp}[DMPs]{Dynamical Movement Primitives}
    \acro{gmm}[GMMs]{Gaussian Mixture Models}
\end{acronym}

\section{Introduction}
\IEEEPARstart{E}{ndovascular} procedures are a rapidly emerging field in medicine. The number of patients treated has constantly increased over the past few decades \cite{durko2018annual}. These procedures increase patient comfort, reduce risks, and improve outcomes compared to traditional open surgery. However, navigation through narrow, fragile, and deformable vessels, using traditional non-steerable catheters and guidewires, requires considerable skill and experience \cite{aggarwal2006virtual}. Steerable catheters and navigation guidance could potentially lower the skill that would be required for percutaneous treatment \cite{culmone2021follow}. Commercial robotic platforms can attest to the robot-assisted trend, such as CorPath™ GRX (Corindus, Waltham, USA), Sensei™ X and Magellan (J\&J robotics, New Brunswick, USA), Amigo™ (Catheter Robotics Inc. Budd Lake, USA), R-One™ (Robocath, Rouen, France) and Niobe™ (Stereotaxis, St. Louis, USA). In the last decades, different research groups have focused their efforts on the development of steerable catheters \cite{da2020challenges,ali2020first, nayar2021design, atlas2022proof}. For example, a proof-of-concept medical robotic platform, composed of a multi-lumen catheter shaft and magnetically actuated microcatheter, was developed in \cite{atlas2022proof}.

In this study, we explore the application of a novel robotic catheterization system, as detailed in \cite{borghesan2020atlas, atlas2022proof}, in the context of \ac{pci} for \ac{cto} treatment. This system employs a robotic catheter, innovatively designed with one to two internal lumens. These lumens serve the critical function of carrying microcatheters with magnetic steering capabilities. A key operational aspect of this system involves the precise navigation of the robotic catheter to an anchor position proximal to the aortic root. Upon reaching this location, the magnetic microcatheters are subsequently deployed into the coronary arteries, setting the stage for the ensuing treatment procedures. The primary objective of this research is to develop a safe, accurate, and robust path planner for the robotic catheter. This path planner is essential for ensuring the robotic catheter's efficient navigation to the aforementioned anchor position near the aortic root. Limited steering capability underscores the need for reliable path planning \cite{favaro2021evolutionary}. However, the complex interaction between the steerable catheter and vessel walls and the deformable property of the vessels makes reliable and real-time path planning a hard problem.

This work presents a robust and accurate path planning framework to improve risk management. Specifically, this framework can reduce the uncertainty in vessel deformation, thereby minimizing tracking errors. The main contributions are:

\begin{itemize}
    \item proposing a novel path planning approach, named \ac{cgail}, which outperforms existing models by offering optimal path planning while adhering to the constraints of robotic catheters. Notably, the \ac{cgail} integrates a \ac{cl} module and a \ac{bc} module, distinguishing it from models like \ac{ppo} + \ac{gail} (\cite{chi_2020, segato2021inverse}), by enabling progressive training in complex environments from demonstrations;
    \item presenting a path planning framework for a motorized steerable catheter, which uniquely considers both the deformable nature of the environment and the dynamic movements of the target, setting it apart from existing path planning methods;
    \item validating the proposed path planner for a motorized steerable catheter in an \textit{in-vitro} setting using a teleoperation control strategy. These experiments underscore the algorithm's feasibility in generating suitable paths that align with the actual steering capability of the catheter, further demonstrating the advantages of the \ac{cgail} model over a traditional centerline-following approach.
\end{itemize}

\section{Related work}

\begin{table*}[t]
    \centering
    \caption{State-of-the-art path planning methods for endovascular catheterization (From 2011 to 2022)}\label{tab:reflist}
    \resizebox{\linewidth}{!}{%
    \begin{tabular}[c]{p{.355\linewidth}p{.035\linewidth}p{.23\linewidth}p{.24\linewidth}p{.08\linewidth}p{.06\linewidth}}
\toprule
\textbf{Reference} & \textbf{Method} & \textbf{Algorithm} & \textbf{Instrument} & \textbf{Environment} & \textbf{Validation} \\
\hline

\cite{Wang_2011} Wang 2011 & NB & Centerline-based tree & Shaped catheter & Rigid & \textit{in-vitro} \\
\cite{zheng20183d} Zheng 2018 & NB & Centerline-based tree & - (Not specified) & Deformable & \textit{in-vitro} \\
\cite{huang2011interactive} Huang 2011 & NB & Depth first search & Guidewire & Rigid & \textit{in-silico} \\
\cite{qian2019towards} Qian 2019, \cite{cho2021image} Cho 2021, \cite{schegg2022automated} Schegg 2022 & NB & Dijkstra & Guidewire & Rigid & \textit{in-vitro} \\
\cite{ravigopal2021automated, ravigopal2022fluoroscopic} Ravigopal 2022 & NB & Modified hybrid A* & Steerable guidewire & Deformable & \textit{ex-vivo} \\
\cite{fagogenis2019autonomous} Fagogenis 2019 & NB & Wall-following & Concentric tube robot & Deformable & \textit{in-vivo} \\
\hline

\cite{fauser2018generalized}, \cite{fauser2019planning}, \cite{fauser2019optimizing} Fauser 2019 & SB & bi-RRT & Catheter / Steerable guidewire & Rigid & \textit{in-silico} \\
\cite{guo2021training} Guo 2021 & SB & RRT & Catheter & Rigid & \textit{in-silico} \\
\hline

\cite{gao2015three} Gao 2015 & OB & Ant colony optimization & Catheter & Rigid & \textit{in-silico} \\
\cite{qi2019kinematic} Qi 2019 & OB & Optimal inverse kinematics & Steerable catheter & Rigid & \textit{in-vitro} \\
\cite{li2021path} Li 2021 & OB & Genetic algorithm & Steerable catheter & Rigid & \textit{in-silico} \\
\hline

\cite{Rafii_Tari_2013}, \cite{rafii2014hierarchical} Rafii-Tari 2014 & LB & GMM, HMM & Shaped catheter, pre-loaded guidewire & Rigid & \textit{in-vitro} \\
\cite{Chi_2018}, \cite{Chi_2018_2}, \cite{chi_2020} Chi 2020 & LB & DMPs, GMMs, GAIL & Shaped catheter, pre-loaded guidewire & Deformable & \textit{in-vitro} \\
\cite{zhao2022surgical} Zhao 2022 & LB & Generative adversarial network & Guidewire & Rigid & \textit{in-vitro} \\
\cite{tibebu2014towards} Tibebu 2014, \cite{you2019automatic} You 2019 & LB & Q-learning, deep Q-learning & Steerable catheter & Rigid & \textit{in-vitro} \\
\cite{behr2019deep} Behr 2019, \cite{karstensen2020autonomous} Karstensen 2020, \cite{kweon2021deep} Kweon 2021 & LB & deep Q-learning, DDPG & Shaped guidewire & Rigid & \textit{in-vitro} \\
\cite{meng2021evaluation} Meng 2021 & LB & Asynchronous advantage actor-critic & Guidewire & Rigid & \textit{in-silico} \\
\cite{karstensen2022learning} Karstensen 2022 & LB & DDPG & Shaped guidewire & Deformable & \textit{ex-vivo} \\
\hline
\textbf{Proposed} & LB & C-GAIL & Steerable catheter & Deformable & \textit{in-vitro} \\
\bottomrule
\multicolumn{6}{l}{\textbf{Acronyms}: NB, Node-Based; SB, Sampling-Based; OB, Optimization-Based; LB, Learning-Based; RRT, Rapidly-exploring Random Tree; GMM, Gaussian Mixture Model;}\\
\multicolumn{6}{l}{HMM, Hidden Markov Model; DMP, Dynamical Movement Primitive; GAIL, Generative Adversarial Imitation Learning; DDPG, Deep Deterministic Policy Gradients.}
    \end{tabular}}
\end{table*}

Over the last decade, several path planning methods for steerable/non-steerable catheters/guidewires have been proposed to assist clinicians. Table~\ref{tab:reflist} summarizes the state-of-the-art from 2011 to 2022, in terms of path planning methodology, type of medical instrument used, type of environment (presence of dynamic changes), and type of validation (\textit{in-silico}, \textit{in-vitro}, \textit{ex-vivo}, \textit{in-vivo}). In the following a very brief description of the main types of planners is given. For a more detailed literature review, please refer to \cite{pore2023autonomous, wu2024review}.

\subsection{Node-Based (NB) Methods}

Node-based algorithms use an information structure to represent the environment map. Studies \cite{Wang_2011} and \cite{zheng20183d} extracted the vessel centerline and built an exploration tree along the centerline. The aim of this method is to keep the tip of the instrument away from the walls. Nevertheless, path exploration inside the information structure is not mentioned in those studies. Graph search strategies such as \ac{dfs} algorithm \cite{huang2011interactive}, Dijkstra algorithm \cite{qian2019towards, cho2021image, schegg2022automated} and A* algorithm \cite{ravigopal2021automated, ravigopal2022fluoroscopic} were employed to generate a path solution in a tubular environment with multi-branches. For movement in the cardiac chamber, a wall-following strategy employing haptic vision was developed in \cite{fagogenis2019autonomous} by Fagogenis \emph{et al.} to keep a certain distance from the heart wall.

\subsection{Sampling-Based (SB) methods}

Sampling-based methods randomly sample in the robot's configuration space or workspace to generate new tree vertices. Then collision-free vertices are connected as tree edges. Fauser \emph{et al.} \cite{fauser2018generalized, fauser2019optimizing, fauser2019planning} introduced a \ac{birrt} method for instruments that follow curvature constrained trajectories in vena cava or aorta. The study in \cite{guo2021training} implemented an improved RRT algorithm for cerebrovascular interventions. The expansion direction of the random tree is a compromise between the new randomly sampled node and the target. This strategy can improve the convergence speed of the algorithm. However, their work did not take into account any kinematic constraints governing catheter movement.

\subsection{Optimization-Based (OB) Methods}

Path planning can be formulated as an optimization problem and solved by numerical solvers. Gao \emph{et al.} \cite{gao2015three} proposed an improved \ac{aco} algorithm to plan an optimal path that also accounted for factors such as catheter diameter, vascular length, diameter, curvature and torsion. Nevertheless, the high computational time with an average value of 12.32s makes it infeasible in real-time scenarios. Qi \emph{et al.} \cite{qi2019kinematic} formulated the path planning as an optimization problem under the inverse kinematics modeling of continuum robots. However, the optimization problem is solved locally without considering long-term cumulative costs. Li \emph{et al.} \cite{li2021path} proposed a fast path planning approach via a local \ac{ga} optimization. The approach is able to account for constraints on the catheter curvature, but the optimization algorithm is based on vessel centerlines that are sensitive to deformations of the anatomical model.

\subsection{Learning-Based (LB) Methods}

Learning-based methods use statistical tools and machine learning algorithms for path planning. Rafii-Tari \emph{et al.} \cite{Rafii_Tari_2013, rafii2014hierarchical} and Chi \emph{et al.} \cite{Chi_2018, Chi_2018_2, chi_2020} proposed \ac{lfd} approaches to optimize trajectories or learn motion primitives using expert demonstrations. Zhao \emph{et al.} proposed a \ac{GAN} framework for real-time path planning and evaluated it in 2D-DSA images \cite{zhao2022surgical}. The work in \cite{you2019automatic, behr2019deep, karstensen2020autonomous, tibebu2014towards, kweon2021deep, meng2021evaluation, karstensen2022learning} developed \ac{rl} approaches to predict a sequence of actions to reach a target. \ac{lfd} methods based on \ac{gail} were adapted into other medical scenarios \cite{pore2021learning, segato2021inverse} because of their ability to compromise between learning the distribution and ensuring the generalization of trajectories. In comparison with the \ac{ppo}+\ac{gail} model employed in neurosurgery as described in \cite{segato2021inverse}, our developed \ac{cgail} network incorporates additional enhancements through the inclusion of both the \ac{bc} and \ac{cl} modules. Moreover, the deformation dynamics in our endovascular intervention setting markedly diverge from those in neurosurgery. Specifically, deformation in our context is induced not solely by the catheter's interaction with the vessel walls but also by the periodic motion associated with the heartbeat. Conversely, neurosurgical procedures typically involve a curvilinear needle trajectory, a simpler navigational challenge than the complex, long, and tortuous routes encountered in blood vessel navigation. These distinctions underscore the specialized challenges of path planning within endovascular interventions relative to neurosurgical applications.

\subsection{Limitations}

Current approaches lack planning capabilities that actually take into account the deformable nature of the environment, even while those studies were verified in a soft environment \cite{zheng20183d, fagogenis2019autonomous, Chi_2018, Chi_2018_2, chi_2020, karstensen2022learning, ravigopal2022fluoroscopic}. Moreover, most of the studies that looked at deformable environments were actually developed for passive, non-steerable instruments \cite{zheng20183d, Chi_2018, Chi_2018_2, chi_2020, karstensen2022learning}. The wall-following algorithm \cite{fagogenis2019autonomous} was only tested on a short path along the inner heart-wall. This approach could be considered efficient if there are few feasible routes to reach the target. However, in scenarios where there are multiple feasible routes, the solution provided by a wall-following algorithm cannot ensure optimality and may cause the catheter to enter other branches along the vessel wall. This algorithm has limited application scenarios. For navigation along vessels, wall-following is not advisable as it could cause the catheter to come into excessive contact with fragile tissue, plaque or calcium that should actually be avoided.

In summary, there is a need for a reliable path planner elaborated in this work that takes into account the deformable nature of the environment and the kinematics of steerable catheters. The paper is built up as follows. Section~III introduces the modeling and path planning methods. Section~IV presents an \textit{in-silico} and \textit{in-vitro} experimental setup, followed by experimental results in Sec.~V. Conclusions and future directions are summarized in Sec.~VI.

\section{Materials and Methods}
\subsection{Moving Agent}

The tip of the catheter is considered as the moving agent. The movement of the catheter is fully determined by the tip under the assumption of \ac{ftl} deployment \cite{culmone2021follow}. While this assumption may not seem very realistic for a catheter with a single bendable segment, it will be shown that it leads to reasonable results.

\begin{figure}[tb]
\centering
\includegraphics[width=\linewidth]{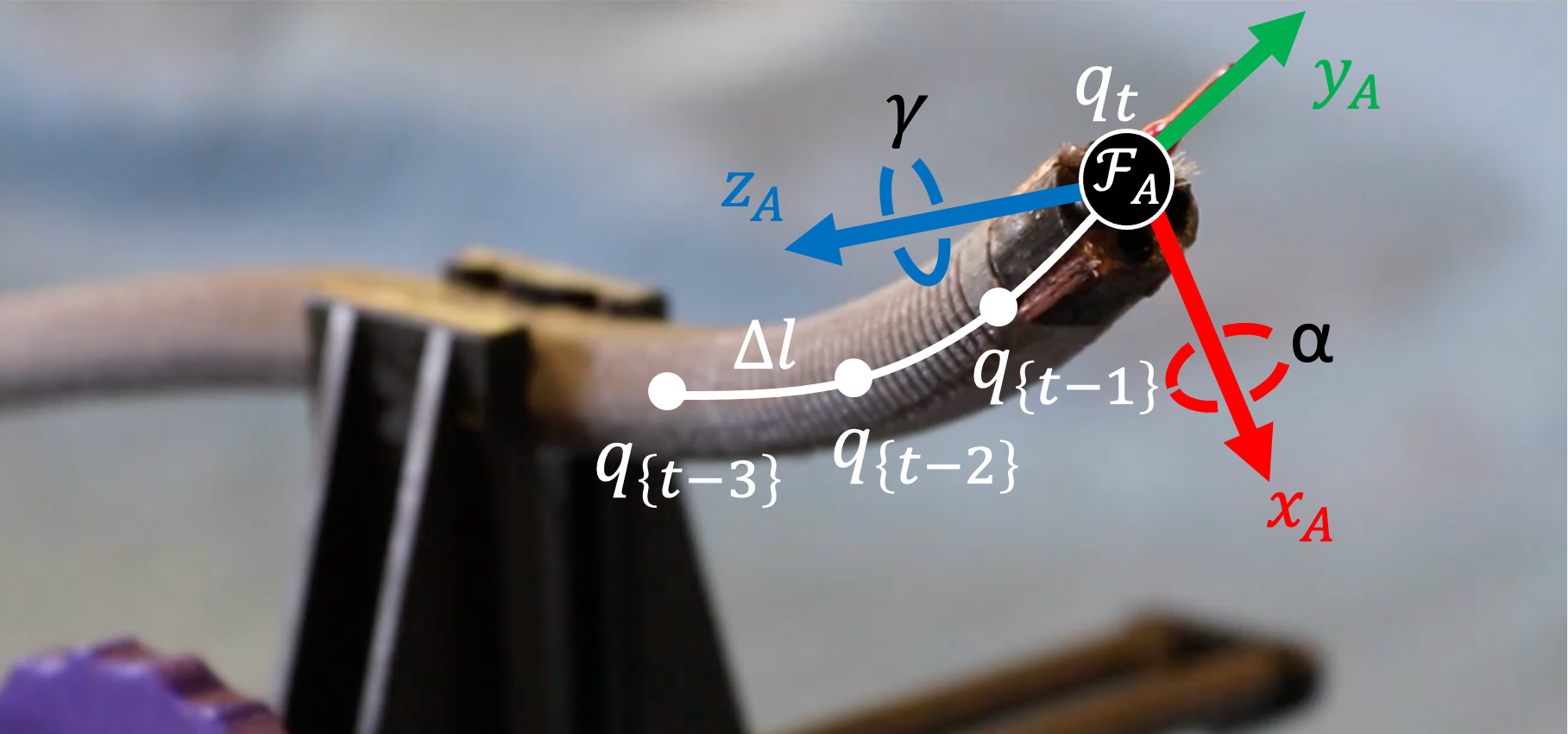}
\caption{Parameterization of a robotic catheter agent: the catheter tip has configuration $\bm{q_t}$ at time $t$. The agent can perform an insertion movement $\Delta_l$ along the $y_A$ axis and can bend with angle $\alpha$ about the $x_A$ axis and with angle $\gamma$ about the $z_A$ axis, respectively, in the tip frame $\mathcal{F}_A$. The catheter segments following the tip adopt the previous configurations sequentially.}\vspace{-10pt}
\label{fig:catheter}
\end{figure}

A fixed coordinate frame $\mathcal{F}_A$ is attached to the tip of the catheter as shown in Fig.~\ref{fig:catheter}. The agent can perform an insertion movement $\Delta_l$ along the $y_A$ axis and can bend with angle $\alpha$ about the $x_A$ axis and with angle $\gamma$ about the $z_A$ axis, respectively. The pose of the agent is determined and updated at each time step by the 3-dimensional continuous action space $\mathcal{A} = [\alpha, \gamma, \Delta_l]$. The pose is defined by the tip's position $\bm{p_t} = [x,y,z]$ and orientation $\bm{r_t} = [\alpha , 0, \gamma]$ in a global frame $\mathcal{F}_0$. Using a transformation matrix, the agent configuration $\bm{q_t}$ can be defined by its pose as below, where the superscript $T$ means transpose.
\begin{equation}
\bm{q_t} = 
\begin{bmatrix}
\bm{R}(\alpha , 0, \gamma) & [x,y,z]^T \\
\bm{0}^T & 1
\end{bmatrix}
\end{equation}

The geometric constraints of the catheter, such as its outer diameter and the length of the distal segment $L$, along with the kinematic constraints, such as the maximum bending angle $\theta_{max}$ and the maximum insertion speed, are considered in the agent's motion. $\theta_{max}^{t}$ is the maximum bending angle the catheter can bend at time $t$, given an insertion $\Delta_l$. The bending angle is within the range of $[-\theta_{max}^{t}, \theta_{max}^{t}]$. This range further depends on the insertion speed $v_t$ and time interval $\Delta_t$ because $\Delta_l=v_t\Delta_t$.
\begin{equation}
\theta_{max}^{t} = \frac{\theta_{max} \Delta_l}{L}
\end{equation}

\subsection{Dynamic Environment}
An effective planner should address the level of uncertainty that is present in this problem. Due to the deformable nature of vessels, pre-planned paths will deviate from the reality. Rigidly following such outdated paths may lead to intense contact with the fragile anatomy. 
A realistic and auto-adaptive simulator to predict vessels' global deformation induced by the catheter's contact and cyclic heartbeat motion was proposed in our previous work \cite{li2022position, li2021development}. The vessel modeling is based on a \ac{pbd} approach. It discretizes an object into a particle system composed of particles. Then it computes the system's time evolution by directly updating particle positions, subject to a set of equality and inequality constraints. In the developed simulator \cite{li2022position}, the deformable property was calibrated according to a stress-strain curve which appropriately depicts the biomechanics properties. Moreover, the heartbeat motion was calibrated according to the averaged annulus displacement from 60 patients with aortic stenosis.

\begin{figure}[tb]
\centering
\includegraphics[width=0.9\linewidth]{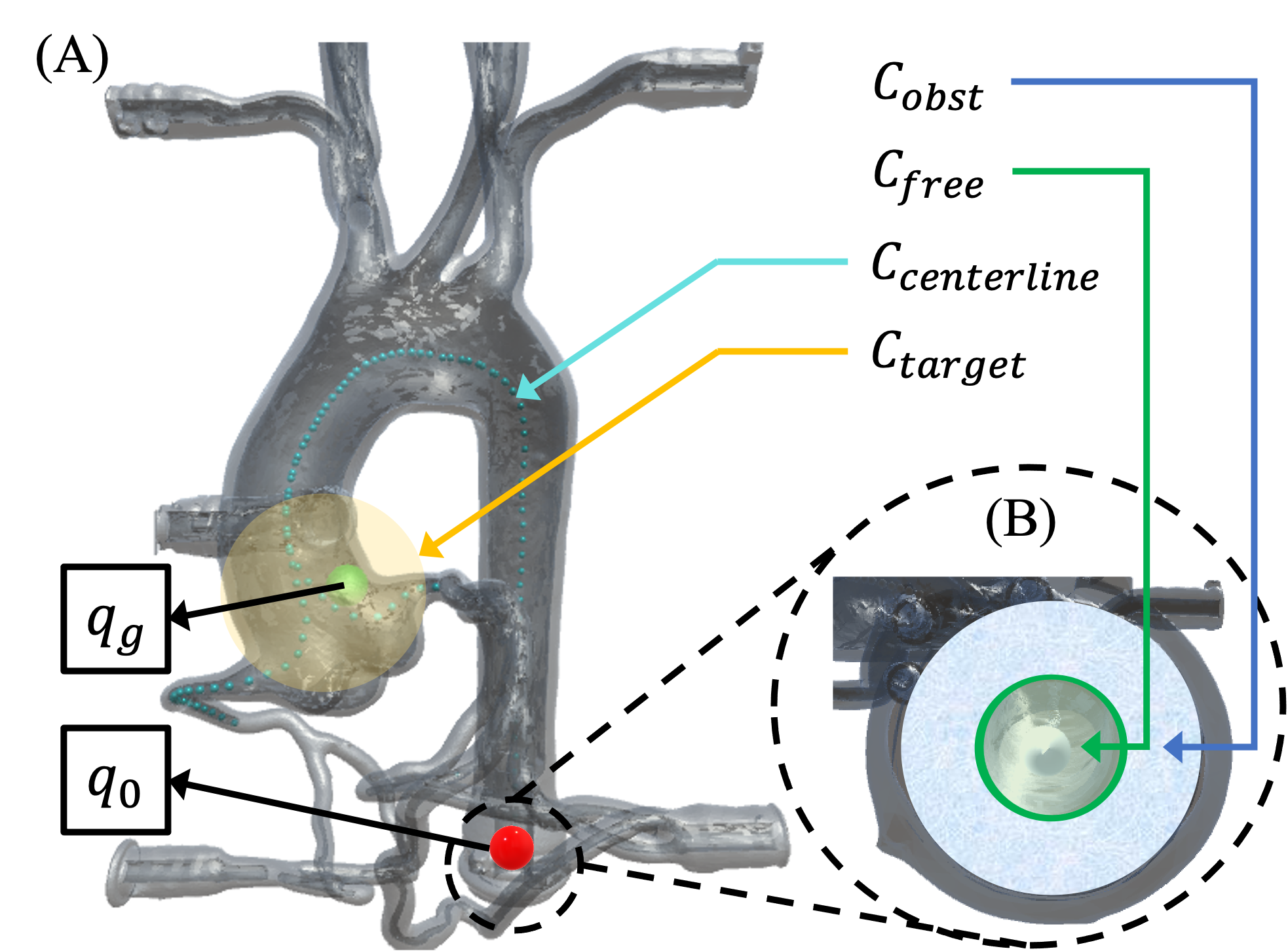}
\caption{The environment is represented by an aortic anatomy, the obstacle space $C_{obst}$, the free space $C_{free}$, the centerline space $C_{centerline}$, and the target space $C_{target}$. The catheter moves from the start configuration $\bm{q_0}$ and proceeds to move to reach the target configuration $\bm{q_g}$. (A) Top view on the aortic model. (B) Cross-sectional view of the open lumen of the descending aorta.}\vspace{-10pt}
\label{fig:areas}
\end{figure}

In this work, for the construction of the 3D dynamic environment (see Fig.~\ref{fig:areas}), let us define:
\begin{itemize}
  \item the ``configuration space" $C_{space}$ as the set of all the possible agent configurations $\bm{q_t}$;
  \item the ``obstacle space" $C_{obst} \subset C_{space}$ that is the space occupied by the vessel wall that limits the area in which the catheter can move; 
  \item the ``free space" $C_{free} \subset C_{space}$ that is the set of all possible agent configurations $\bm{q_t}$ within the aorta lumen without collisions with other objects;
  \item the ``centerline space" $C_{centerline}$ that is the shortest path computed via the Voronoi Diagram from the descending aorta to the left and right coronaries;
  \item the ``target space" $C_{target} \subset C_{space}$ that is the volume where the target configuration $\bm{q_g}$ can locate. Once the delivery catheter reaches the target space, a micro-catheter can be inserted from a channel of the delivery catheter \cite{atlas2022proof}. $\bm{q_g}$ changes randomly at every learning episode within $C_{target}$. Importantly, the target moves in concert with the deformation of the vessels, which is induced by the catheter's contact and the cyclic motion of the heartbeat. This movement is achieved by linking the target with one of the particles within the \ac{pbd} system; 
  \item the agent start configuration $\bm{q_0}$ that is located in the descending aorta.
\end{itemize}

\subsection{Path Planning}

The path planning problem can be described as: the agent has to find an admissible set of configurations $Q_t = \{ \bm{q_0} , ... , \bm{q_g} \}$ to move from a start position $\bm{p_0} \in \bm{q_0}$ to a target position $\bm{p_g}\in \bm{q_g}$.
The target position is reached when the distance between the agent and the target is smaller than a distance threshold $\epsilon$.

The state of the agent consists of its pose, that can be changed through the actions of rotation and insertion $\mathcal{A} = [\alpha, \gamma, \Delta_l]$. In training, the agent learns to maximize the reward function by taking actions according to its policy $\tau$, expressed in the paragraph \textit{Reward Function} and by observing its interaction with the environment, described in the subsequent paragraph \textit{Observations}.

\subsubsection{Reward Function}
The reward function $R(\tau)=r_t$ associated with each time step $t$ is designed to optimize the path according to a combination of multiple criteria: the number of steps, the number of collisions, reaching the target position, passing through the centerline waypoints, bending angle. The reward $r_t$ consists of two main parts: $r_{end}$, a reward added at the end of a learning episode; $r_{in}$, a relatively small reward added at each step during a learning episode. The reward $r_t$ is expressed in (\ref{eq:rt})-(\ref{eq:rin}).

\begin{equation}\label{eq:rt}
r_t = r_{end} + r_{in} \,
\end{equation}
\begin{equation}
r_{end} =
    \begin{cases}
     r_{obst} \quad \textrm{if} \quad \bm{q_t} \in C_{obst} \\
     r_{exit} \quad \textrm{if} \quad \bm{q_t} \notin C_{free} \, \textrm{and} \, \bm{q_t} \notin C_{obst} \\
     r_{target} \quad \textrm{if} \quad ||\bm{p_t} - \bm{p_g}|| < \epsilon
    \end{cases} \,
\end{equation}
\begin{equation}\label{eq:rin}
    r_{in}=r_{step}+r_{centerline}+r_{bending}
\end{equation}
\begin{itemize}
    \item $r_{obst}$ is a negative reward that is given if a collision between the catheter tip and vessel walls is detected. We only take that collision into account because it has a higher risk during navigation. An episode terminates when a non-minor collision occurs;
    \item $r_{exit}$ is a negative reward when the agent tries to exit from the open lumen down the descending aorta;
    \item $r_{target}$ is a positive reward given to the agent when it reaches the target;
    \item $r_{step}$ is a negative reward given at each time step. It is set to keep the total number of steps of the trajectory as small as possible;
    \item $r_{centerline}$ is a positive reward if the agent reaches a waypoint in $C_{centerline}$;
    \item $r_{bending}$ is a positive reward that is given when the bending action is bigger than a threshold. This reward was introduced to overcome the tendency of the network to avoid producing actions near the catheter's maximum bending angle. To pass tortuous areas, the maximal bending range is often needed to be able to pass. 
\end{itemize}

The values of the reward function parameters obtained with an empirical method are summarized in Table~\ref{tab:rewards}. All rewards are set within the interval of [-1,1]. The minimum value of ``-1" is assigned to prohibited behaviors such as violent collisions with the vessel walls. The maximum value of ``1" is assigned to reaching the target, which is the agent's task. The other rewards are chosen based on their frequency. For instance, since $r_{step}$ occurs very often, if it is not small enough, it can lead to a large cumulative reward. Similarly, the bending reward also has the potential to lead to a large cumulative reward. In contrast, the number of centerline points is relatively small, with only around 100 points, and not all of them are reachable if the catheter constraints are met. Hence, this positive reward is set slightly larger.

\begin{table}[tb]
\centering
\caption{Values of the reward function parameters}\label{tab:rewards}
\begin{tabular}{p{.15\linewidth}p{.15\linewidth}p{.2\linewidth}p{.15\linewidth}}
\toprule
\textbf{Reward} & \textbf{Value} & \textbf{Reward} & \textbf{Value} \\ \hline\noalign{\smallskip}
$r_{obst}$ & -1 & $r_{step}$ & -1e-5 \\
$r_{exit}$ & -1 & $r_{centerline}$ & +0.05 \\
$r_{target}$ & +1 & $r_{bending}$ & +1e-5 \\
\bottomrule
\end{tabular}
\end{table}

\subsubsection{Observations} At every step, the agent collects observations $o_t$, which are composed of:
\begin{itemize}
    \item the agent configuration $\bm{q_t}$;
    \item the normalized distance from the agent to the target $u = \frac{||\bm{p_g} - \bm{p_t}||}{d_{max}}$, where $d_{max}$ is the maximum distance between the agent and the target when $\bm{q_t} \in C_{free}$;
    \item the direction from the agent to the target $\bm{v} = \bm{p_g} - \bm{p_t}$;
    \item a set of raycast observations $o_{ray}$. Each raycast detects the presence of the aortic wall along its direction within the ray length.
\end{itemize}

\subsubsection{C-GAIL network}
\begin{figure}[tb]
\centering
\includegraphics[width=\linewidth]{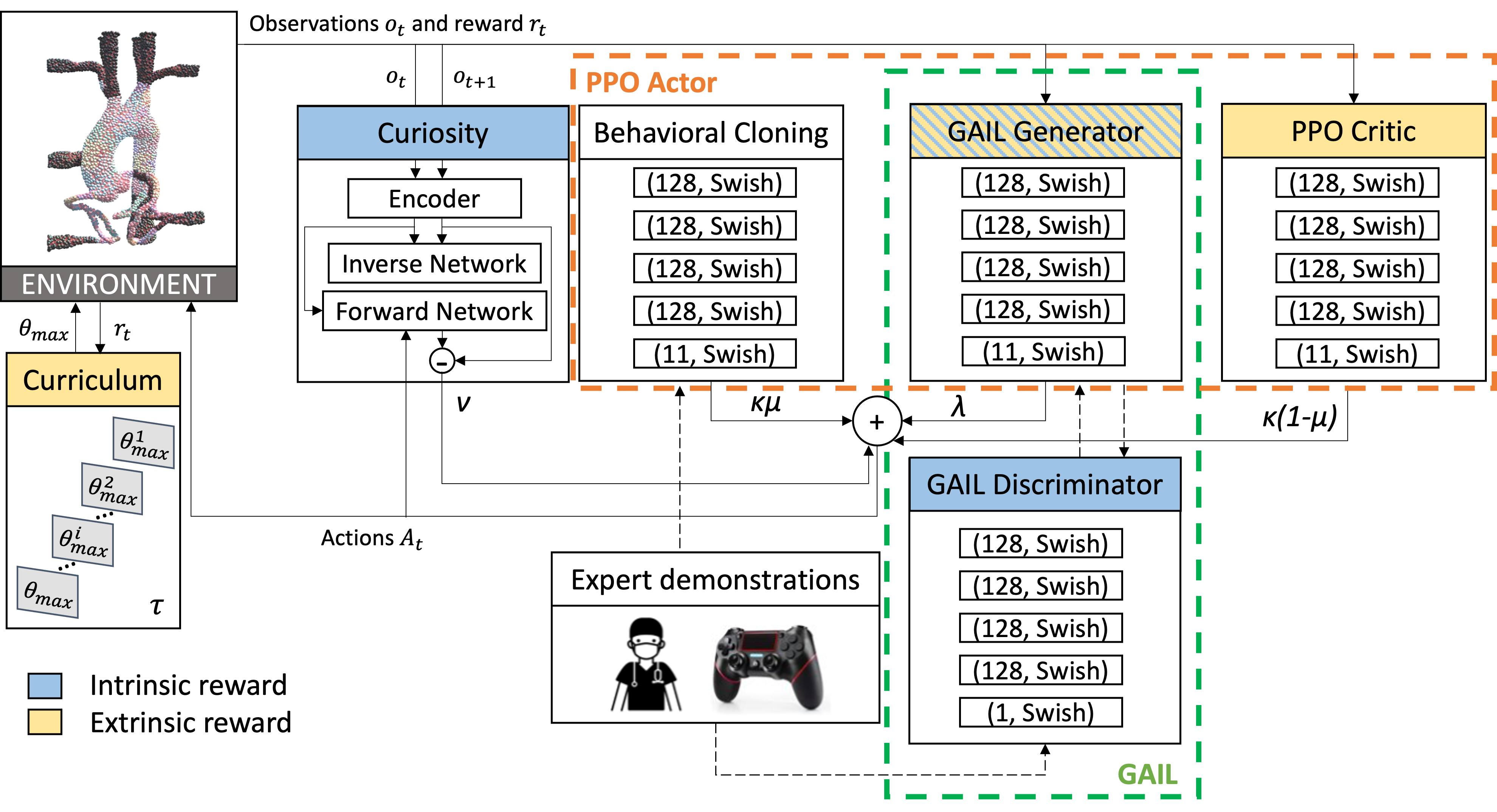}
\caption{The proposed \ac{cgail} network architecture. The extrinsic reward signal considers the reward given by interacting with the environment, such as curriculum and PPO modules. The intrinsic reward has a policy that considers other factors, and it is defined inside the learning algorithm: for \ac{gail} about the similarity of the path with respect to the expert demonstrations, for curiosity about the difference between the predicted and the actual path.}\vspace{-15pt}
\label{fig:network}
\end{figure}

The new network proposed in this paper, named \ac{cgail} network, is built around the combination of the principles of \ac{lfd} and \ac{rl}  (see Fig.~\ref{fig:network}). The \ac{lfd} component is realized through \ac{bc} and \ac{gail} networks. 
The \ac{ppo} network has two parts: the actor network provides an action given the observation; the critic network that evaluates the actor network using the extrinsic reward and suggests modification according to a gradient ascent policy. The actor network updates its actions according to three policies: the \ac{bc} policy, \ac{ppo} policy, and the \ac{gail} generator policy. The curiosity module \cite{pathak17a} acts as an intrinsic reward signal that enables the agent to explore its environment in novel states to help escape local minima of the policy function. The curiosity module contains an inverse and forward network. The inverse network predicts the action between observations, while the forward network predicts the next encoded observation. The difference between the predicted and actual encoded observations is defined as the loss of the forward network. Therefore, through curiosity-driven exploration, the agent can predict the outcome of its actions and acquire skills that may be valuable in the future. The curriculum learning module acts on the environment by progressively adding complexity during the training \cite{narvekar2016source}. The curriculum learning module optimizes the bending angle while respecting the reachable bending range of the catheter. Specifically, in curriculum learning, the learning progress is measured through the reward function and once the agent performance improves, $\theta_{max}$ is decreased for the next level of learning. Finally, the agent is able to try actions in different reachable bending ranges and obtain globally optimal paths.

The proposed network can be represented as a set of modules with weights, that define the contribution of each module to the loss function. These modules interact with the environment and with each other. The goal is to come up with an optimal combination of the strengths of the different modules such that a performance is achieved that exceeds those of the offered demonstrations. The training is based on the linear combination of the respective \ac{rl} and \ac{lfd} losses:
\begin{equation}\label{eq:losses}
    \mathcal{L} = \kappa(1-\mu) {\mathcal{L}}_{PPO} + \lambda {\mathcal{L}}_{GAIL} + \kappa\mu {\mathcal{L}}_{BC} + \nu {\mathcal{L}}_{curiosity}
\end{equation}
where $\mathcal{L}_{PPO}$ is the \ac{ppo} critic network loss (see its definition in \cite{schulman2017proximal}),  $\mathcal{L}_{GAIL}$ is the \ac{gail} loss \cite{ho2016generative},  $\mathcal{L}_{BC}$ is the \ac{bc} loss \cite{torabi2018behavioral},  $\mathcal{L}_{curiosity}$ is the curiosity loss \cite{pathak17a}.  The weight $\mu$ indicates the degree to which we prioritize the influence of \ac{bc} over the policy relative to \ac{ppo}, with a higher weight indicating a higher learning rate of imitation from demonstrations and a lower weight indicating more operations attempting to maximize reward rather than imitating. The following weights were found empirically to work well: $\kappa = 0.2$, $\lambda = 0.8$, $\mu = 0.7$, $\nu=0.02$. The GAIL is given the highest weight because the method combines two paradigms and reward types, intrinsic and extrinsic. The selection process for these weights involved a systematic evaluation. Initially, a wide array of weight combinations underwent testing in preliminary trials to pinpoint those substantially impacting model performance. This was followed by narrowing our focus to specific ranges surrounding these effective values. The definitive weights were selected based on their consistent contribution to enhanced performance metrics, detailed in Sec.~\ref{sec:metrics}.

\begin{figure*}[tb]
\centering
\includegraphics[width=\linewidth]{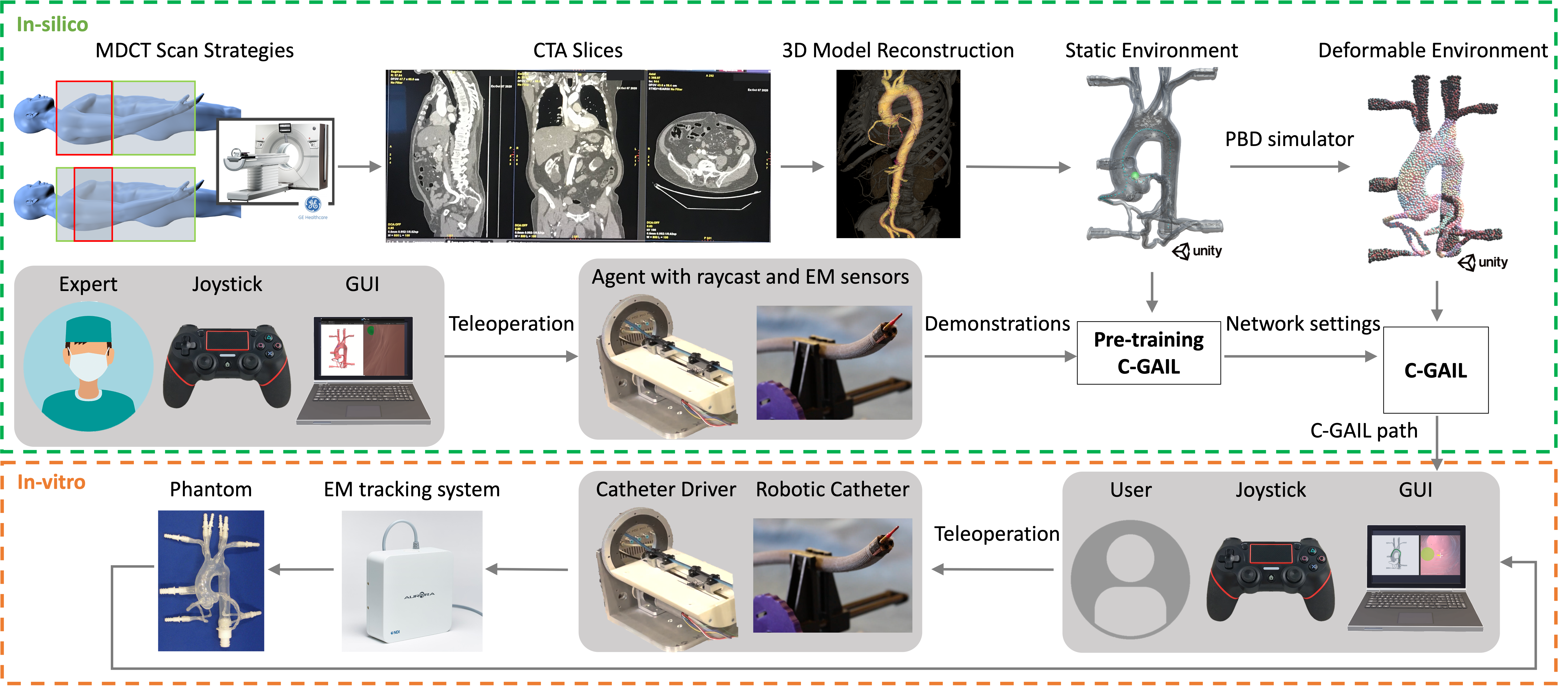}
\caption{Workflow for simulation and \textit{in-vitro} user study. First, a 3D model is reconstructed from CTA images for a specific patient, and a deformable environment is built. Next, based on expert demonstrations, the C-GAIL network is trained to provide an optimal path. This path is then rendered through a GUI for the \textit{in-vitro} experiments and serves as path guidance for users.}\vspace{-15pt}
\label{fig:workflow}
\end{figure*}

\subsection{Clinical Workflow for the Proposed Path Planner}

Regarding the clinical translation of this work, the anticipated clinical workflow is designed as follows. 
\begin{itemize}
    \item Firstly, a 3D mesh model is reconstructed from the pre-operative \ac{cta} images of a specific patient using the \textit{AW server} (GE Healthcare). Based on this 3D model, a patient-specific deformable environment is built using the \ac{pbd} simulator \cite{li2022position}. In this simulator, \ac{pso} is employed to derive the optimal \ac{pbd} parameters that accurately replicate the stress-strain curve of blood vessels and to simulate the vessels' movements resulting from heartbeat. The calibration of heartbeat movements is based on the average annulus displacement derived from 60 patients with aortic stenosis. Furthermore, should data on the specific properties of a patient's blood vessels be available, the simulator can be tailored to reflect patient-specific characteristics.
    \item Then, the proposed \ac{cgail} network undergoes training with expert-provided, patient-specific demonstrations, which may encompass approximately 70 instances. This model can either be trained anew for an individual patient or adapted from an existing model to accommodate minor anatomical differences. The training process in the \ac{pbd} simulator is expected to take approximately six hours.
    \item Given the desired target configuration, the proposed \ac{cgail} path planner gives a set of configurations to move from a starting position to the target and outputs an optimal path for the catheter.
    \item (optional) A patient-specific phantom is manufactured, and clinicians can teleoperate the robotic catheter to perform \textit{in-vitro} experiments. The \ac{cgail} path is rendered through a \ac{gui}, serving as path guidance for the clinicians. This enables clinicians to acquire experience prior to the actual interventions, thereby reducing the likelihood of encountering unforeseen situations.
    \item For \textit{in-vivo}, \textit{ex-vivo}, or clinical practice, the \ac{cgail} path is rendered through a \ac{gui} and serves as path guidance for the clinicians. The clinicians can teleoperate the robotic catheter to reach the clinical target site by following the \ac{cgail} path.
\end{itemize}

\section{Experimental Setup}

The proposed workflow for the simulation and \textit{in-vitro} user study is illustrated in Fig.~\ref{fig:workflow}. Firstly, a 3D mesh model is reconstructed from the \ac{cta} images of a specific patient, and a deformable environment is built using the \ac{pbd} approach \cite{li2022position}. Then, the proposed \ac{cgail} network is trained using demonstrations provided by experts, which outputs an optimal path for the catheter. The \ac{cgail} network utilizes expert demonstrations to learn and transfer expert experience, which can provide path references for non-expert users and even autonomous catheters.

In addition to evaluating the path planning algorithm based on \ac{cgail}, this paper also includes control experiments (path following experiments) to verify the feasibility of executing the planned path. The \textit{in-vitro} experiments used human-in-the-loop teleoperation control to guide the catheter along the planned path. It is worth noting that while control strategies are not the focus of this paper, they were merely included in the experiments to evaluate the feasibility and performance of the path planner. Teleoperation is more commonly used and easier to implement in practice, making it the method of choice for the \textit{in-vitro} experiments. It is important to note that any control algorithm could theoretically be used to execute the path planned in this paper.

\subsection{\textit{In-silico} Path Planning Setup}
\subsubsection{Hardware Specification}
Experiments are carried out on a computer equipped with an Intel(R) Core(TM) i9-9900KF CPU @3.60GHz 3.60 GHz processor and 32.0 GB RAM, with an NVIDIA GeForce RTX2080Ti GPU card.

\subsubsection{Experimental Protocol}
 The purpose of the \textit{in-silico} experiments is to validate that, given the curvature constraints of the robotic catheter, the paths obtained by the proposed \ac{cgail} network have better performance than the state-of-the-art GAIL approach in a static environment and are also capable to operate in a complex dynamic environment.

Training and testing were conducted in a single aortic model. For the training, 70 demonstrations were recorded using a joystick by one expert user, who performed catheter navigation through vessels in the aortic model for a \ac{pci}. The ``expert user" mentioned pertains to an individual proficient with both joystick operation and this specific simulator. Future research endeavors will focus on training cardiologists, who possess at least five years of endovascular procedure experience, in the operation of this simulator system. Subsequently, we plan to collect their demonstrations to further enrich our study. The number of experiments for comparison was set to 100. Different start configurations in the descending aorta and possible target positions were chosen to make the training more generally valid. There are three possible start configurations distributed within 5.3cm along the descending aorta. There are five target positions distributed in a spherical space with a radius of 3cm near the coronaries and the vessel walls. Furthermore, the target moves in response to vessel deformations, which are triggered by the catheter's contact and the periodic motion of the heartbeat. The introduction of variability into the simulation model enriches the training data with diverse configurations. While such variability may initially impede convergence, it has the potential to significantly improve the network model's robustness and generalizability, as discussed in \cite{xiao2022multigoal}. The robustness will be assessed through performance metrics, to be detailed in Section~\ref{sec:metrics}, with a particular focus on success rate and targeting error.

The \ac{cgail} network training was carried out in two phases:  pre-training in a rigid, computationally less heavy environment, where the tuning of the network parameter settings is performed; re-training in a deformable environment, starting with the previously tuned network settings. Pre-training helps reduce the time required to obtain network settings with empirical parameters and compare the differences in agent behavior under different settings. The training parameters for the \ac{cgail} is presented in Table~\ref{tab:parameters}. 

The work from Chi \emph{et al.} \cite{chi_2020} is used as a reference to compare with. It's important to note that we do not directly adopt the parameters from Chi \emph{et al.}'s work, as we are addressing different robotic platforms. Instead, to keep consistency in comparison, we set the same training parameters for the state-of-the-art \ac{gail} and our proposed \ac{cgail}. The \ac{gail} network architecture of \cite{chi_2020} is adopted with minor modifications: each layer has 64 units and a Swish activation function \cite{ramachandran2017searching}. The \ac{gail} network undergoes training with an additional set of 70 demonstrations, recorded by the same expert user and within the same simulation environment as utilized for the \ac{cgail} training process. The prior studies by Chi \emph{et al.} \cite{Chi_2018, Chi_2018_2} introduced an \ac{lfd} method based on \ac{dmp} and \ac{gmm}. Later, they improved the \ac{rl} part by including model-free \ac{gail} loss. Therefore, their \ac{gail} network model was selected for comparison in our study. Other methodologies applied in deformable environments, as summarized in Table~\ref{tab:reflist}, are primarily designed for passive, non-steerable instruments. The wall-following technique described in \cite{fagogenis2019autonomous} is not apt for navigating through vessels with multiple branches. The hybrid A* approach \cite{ravigopal2022fluoroscopic} seeks the optimal path in a 2D plane, whereas the true 3D optimal path is determined using RRT*. However, RRT* may not be ideal in tightly narrow and tortuous vessels due to its potential to produce sub-optimal paths that run dangerously close to vessel walls.

\begin{table}[tb]
\centering
    \caption{Training parameters for  \ac{cgail} and GAIL.}
    \label{tab:parameters}
\begin{tabular}{p{.3\linewidth}p{.1\linewidth}p{.3\linewidth}p{.1\linewidth}} 
\toprule
\textbf{Parameter} & \textbf{Value} & \textbf{Parameter} & \textbf{Value}\\
\hline\noalign{\smallskip}
PPO beta & 5.0e-4 & PPO gamma & 0.99 \\
max steps & 5.0e5 & GAIL gamma & 0.99 \\
batch size & 1024 & buffer size & 10240\\
\bottomrule
\multicolumn{4}{l}{All parameters are defined in the Unity ML-Agents Toolkit \cite{juliani2018unity}.}
\end{tabular}
\end{table}

\subsection{\textit{In-vitro} Validation Setup}
\subsubsection{Hardware Specification} 

\begin{figure}[tb]
\centering
\includegraphics[width=\linewidth]{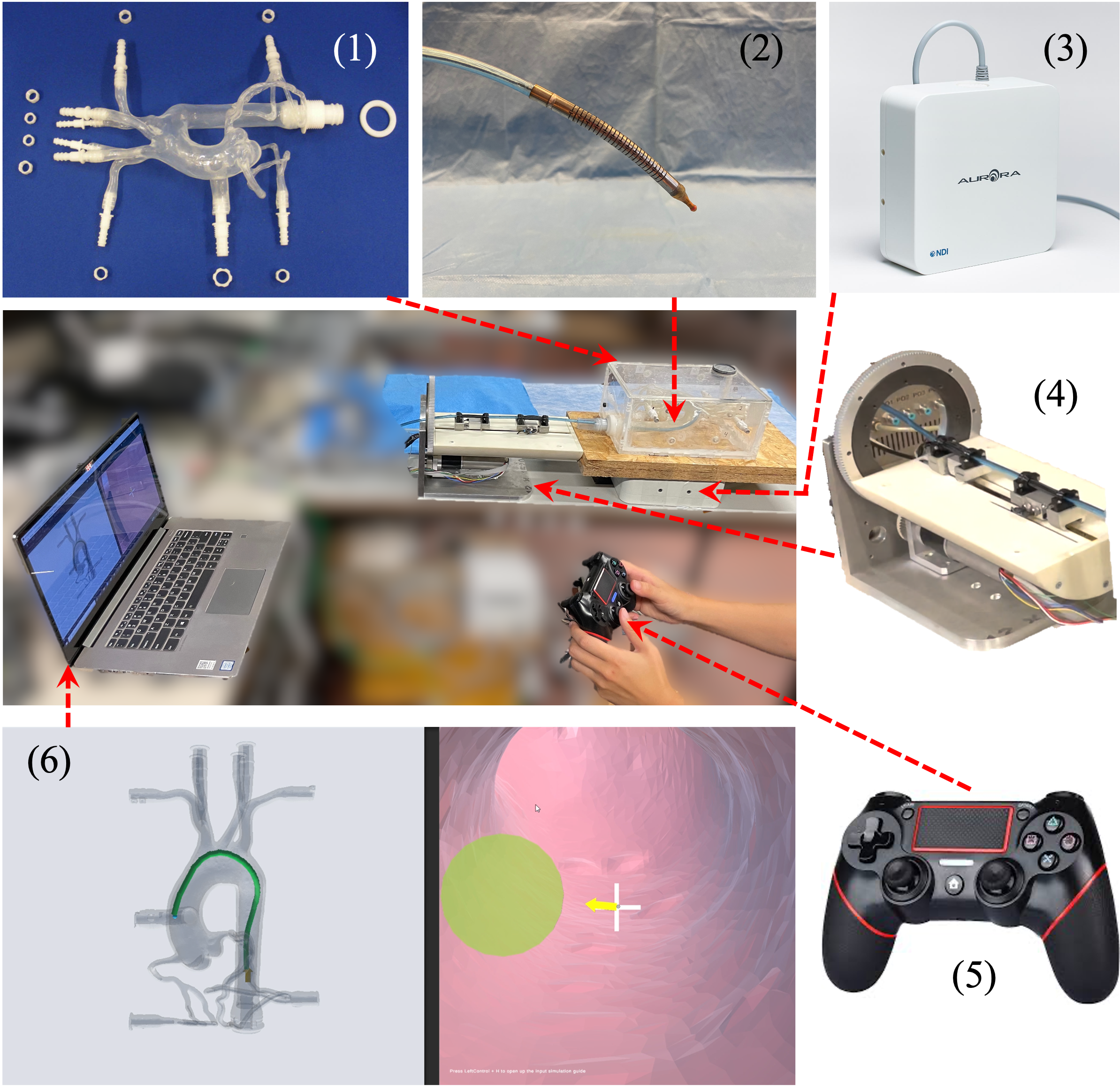}
\caption{\textit{In-vitro} experimental setup to validate the proposed path planning approach: (1) silicone aortic phantom; (2) robotic catheter; (3) Aurora EM field generator; (4) sleeve-based catheter driver; (5) wireless controller; (6) the GUI as visual feedback.}\vspace{-15pt}
\label{fig:setup}
\end{figure}

The \textit{in-vitro} experimental setup, including the following devices, is depicted in Fig.~\ref{fig:setup}.

\paragraph{phantom} experiments were performed in a transparent, deformable silicone aortic phantom (T-S-N-002, Elastrat Sarl, Geneva, Switzerland).

\paragraph{robotic catheter} the custom-made robotic catheter is fabricated out of Nitinol using metal laser cutting technology and is actuated by four integrated \ac{pam} \cite{devreker2015fluidic, wu2021hysteresis}. The outer diameter of the catheter is 7mm, and the length is 900mm. The distal Nitinol segment is 75mm long and includes a 50mm long steerable distal segment. This steerable distal segment has 2-\ac{dof} bending motion by concurrently controlling two antagonistic pairs of the \ac{pam}. The maximum bending angle is $90^{\circ}$. 

\paragraph{electromagnetic field generator} an \ac{em} field generator (Northern Digital Inc., Waterloo, Canada) is placed beneath the phantom. A 6-\ac{dof} \ac{em} sensor (Northern Digital Inc., Waterloo, Canada) is embedded at the tip of the catheter to track its pose. 

\paragraph{catheter driver} the robotic catheter driver system is described in \cite{omar2023force}. The sleeve-based catheter driver has two pneumatically actuated grippers that grasp the catheter alternately and insert the catheter in a relay fashion. The maximum insertion (and retraction) speed is set as 5mm/s. 

\paragraph{joystick} the teleoperated catheter insertion and bending are realized through velocity control of the catheter driver and pressure control of the \ac{pam}, respectively, using a wireless controller (Yues, Dublin, Ireland). 

\paragraph{GUI} the path planning is visualized by a \ac{gui}, which includes an external projected view showing the aorta, path, and pose of the catheter tip, and an internal view showing the next waypoint and suggested bend direction. Guided by the internal view, users are advised on the optimal direction and degree of bend needed from the catheter tip's current position to achieve precise positioning.

\subsubsection{Experimental Protocol}
The objective of the \textit{in-vitro} experiments is to validate whether the path obtained from the proposed \ac{cgail} path planner is more compatible with the actual steering capability of the catheter. Therefore, it can also be verified whether this path can better help the user steer the catheter accurately to the target. 

For the \textit{in-vitro} comparison with \ac{cgail}, we chose the traditional centerline-following approach. The centerline path is widely used in interventions \cite{sobocinski2013benefits, ramcharitar2008technology}, whereas GAIL approach (which was served as a comparison in the \textit{in-silico} experiments) has not yet been validated in clinical practice. Moreover, both GAIL and \ac{cgail} are \ac{rl} approaches, which have limited interpretability. Therefore, it may be suggested to use traditional approaches in parallel with learning-based approaches in order to provide a more transparent and explainable analysis. By comparing \ac{cgail} with the traditional centerline approach, a more interpretable comparison can be achieved.

User-involved control experiments were carried out. A single participant with an engineering background and prior experience with robotic catheters was involved. This individual had not participated in the \textit{in-silico} experiments and was distinct from the expert user who provided the demonstrations for network training. In each trial, the user teleoperates with the robotic catheter, tries to pass through each waypoint, and finally reaches the target. During experiments, the user is instructed not to directly observe the transparent phantom. The user is instead required to rely on visual feedback provided by the \ac{gui} displayed on a conventional 2D monitor for his (her) teleoperation actions. The \ac{gui} provides critical feedback to users. It displays the catheter's tip pose, captured through \ac{em} tracking. By visualizing this alongside the rendered phantom, users can effectively gauge their proximity to the vessel walls. This feature is instrumental in preventing collisions or applying excessive force against the vessels. The maximum operation time is set to 3 minutes. If the maximum operation time is exceeded and the target is still not reached, the experiment will be forcibly stopped and be regarded as a failure. Setting this stopping criterion can avoid the following two cases: 1) The user rushes to finish the experiment without considering the performance; 2) The user tries excessively to get better performance while the procedure is very time-consuming.

The user's performance was compared in two scenarios: with the \ac{cgail} assistance and centerline assistance. The waypoints from $C_{centerline}$ were utilized in the centerline-following approach. The \ac{cgail} network was trained using different targets and obtained a path to reach an unseen target (that was not used during training). To largely eliminate the impact of the user's learning curve, the user is required to be pre-trained. Only when the learning curve converges to a plateau can we assume that the user has stable catheter manipulation abilities. The user pre-training experiments are repeated ten times. After that, control experiments with the same \ac{cgail} path are conducted ten times. Similarly, the pre-training experiments with the centerline and the control experiments with the same centerline are repeated ten times each. The experiments with the \ac{cgail} path are first performed to eliminate experience learned from the other scenario.

\subsection{Performance Metrics}\label{sec:metrics}

To evaluate clinical performance during endovascular procedures, clinicians typically utilize a range of benchmarks, such as patient outcomes, procedural success, and safety measures \cite{saricilar2021evaluation}. Engineering studies \cite{duran2015kinematics, mazomenos2016catheter} have focused on objectively determining cardiologists' proficiency through catheter kinematics analysis. Building upon these established evaluation approaches, our research integrates specific performance metrics to thoroughly quantify the performance of \textit{in-silico} and \textit{in-vitro} experimental results. The statistically significant difference between the proposed method and others is evaluated via the Kruskal-Wallis test in this work, with a significance level of 0.05.

\subsubsection{Success Rate (\texorpdfstring{$\delta$}{TEXT})} Success rate represents the percentage of successes among the total number of attempts.
\begin{equation}
    \delta = n_s / n
\end{equation}
where $n_s$ is the number of successes to reach the target within a given time, and $n$ is the number of attempts.

\subsubsection{Timesteps (\texorpdfstring{$T_s$}{TEXT})} For \textit{in-silico} experiments, at each time step, the agent receives observations, takes an action according to its policy and receives rewards. $T_s$ is defined as the number of time steps moving from the start configuration to the target. 
\begin{equation}
    T_s=N_g - N_0 + 1
\end{equation}
where $N_0$  and $N_g$ are the first and last time step, respectively.

\subsubsection{Duration (\texorpdfstring{$T$}{TEXT})} Duration is the length of time a single experiment lasts from start to stop. An experiment stops when the target or the maximum duration has been reached.
\begin{equation}
    T=t_g-t_0
\end{equation}
where $t_0$ and $t_g$ are the first and last timestamp, respectively. The total duration of the procedure assumes significance not only in terms of its impact on patient safety but also in relation to patient comfort and recovery. Prolonged procedures may lead to increased patient discomfort and extended recovery periods, making duration a valuable indicator of procedure performance, particularly with regard to safety. Moreover, extended durations place a burden on hospital resources, tying up catheterization labs and medical personnel and potentially escalating operational expenses.

\subsubsection{Tracking Error (\texorpdfstring{$T_r$}{TEXT})}
Tracking error is the distance between the \textit{in-vitro} trajectory obtained from the EM sensor and the desired path. The desired path is obtained through path planning, followed by B-spline curve fitting and resampling. This criterion is used to evaluate the path following capability.
\begin{equation}
    T_r(t)=||\bm{p^d_t}-\bm{p_t}||
\end{equation}
where $\bm{p^d_t}$ is the desired position and $\bm{p_t}$ is the actual position. Tracking error is employed as a safety metric, with lower tracking errors denoting a trajectory that closely matches the intended path, thereby indicating a lower likelihood of complications, including vascular damage.

\subsubsection{Targeting Error (\texorpdfstring{$T_a$}{TEXT})}
Targeting error is the minimum distance to the target along the trajectory. This criterion is used to evaluate the accuracy when reaching the target.
\begin{equation}
    T_a=\min||\bm{q_g}-\bm{p_t}||
\end{equation}
Targeting error is utilized as an indicator of procedural technical success, where reduced targeting error represents precise device placement.

\section{Results and Discussion}
\begin{figure}[tb]
\centering
\includegraphics[width=\linewidth]{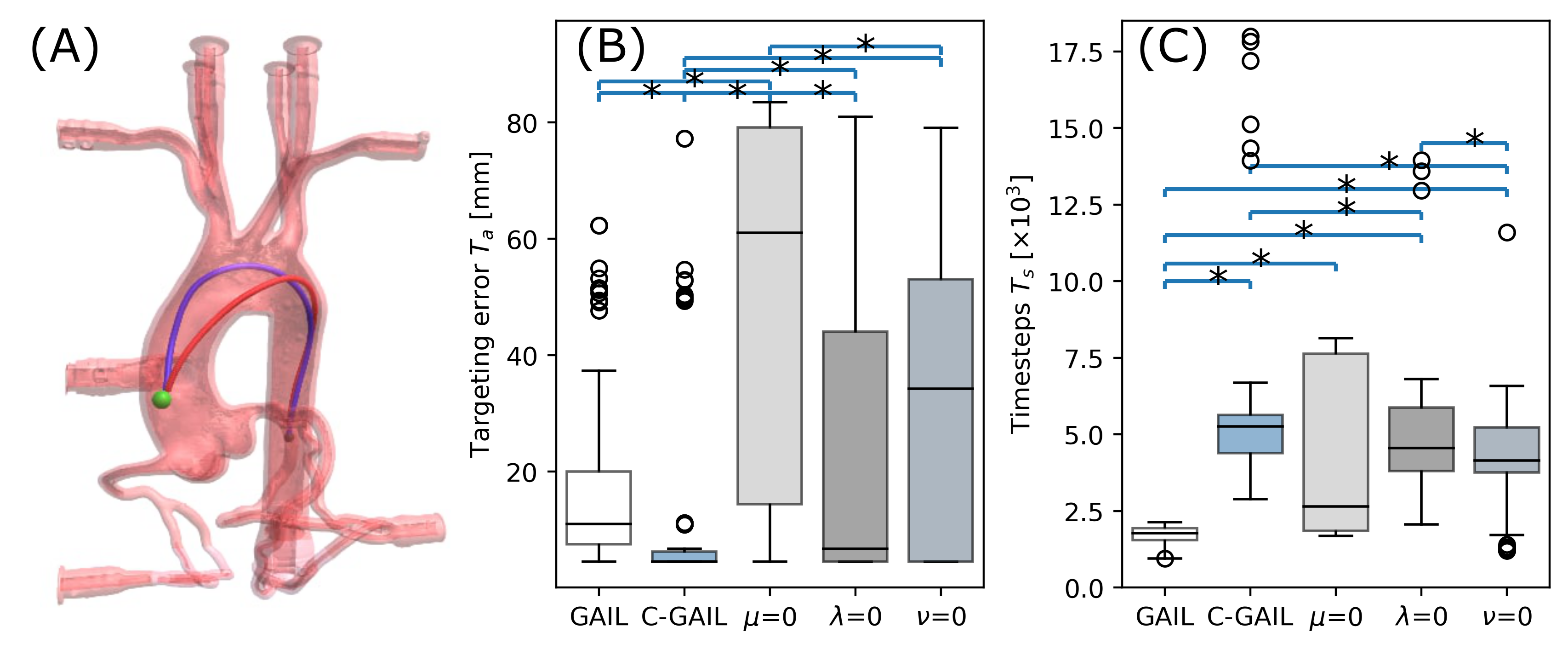}
\caption{\textit{In-silico} performance comparison between the proposed \ac{cgail} network (purple) and the state-of-the-art GAIL network \cite{chi_2020} (red), with respect to (A) example of trajectories, (B) targeting error $T_a$, and (C) timesteps $T_s$. An ablation study of the \ac{cgail} is presented when $\mu=0$, $\lambda=0$ or $\nu=0$. ($*, p<0.05$ using Kruskal-Wallis test)}
\label{fig:comp_path}
\end{figure}
\begin{figure}[tb]
\centering
\includegraphics[width=\linewidth]{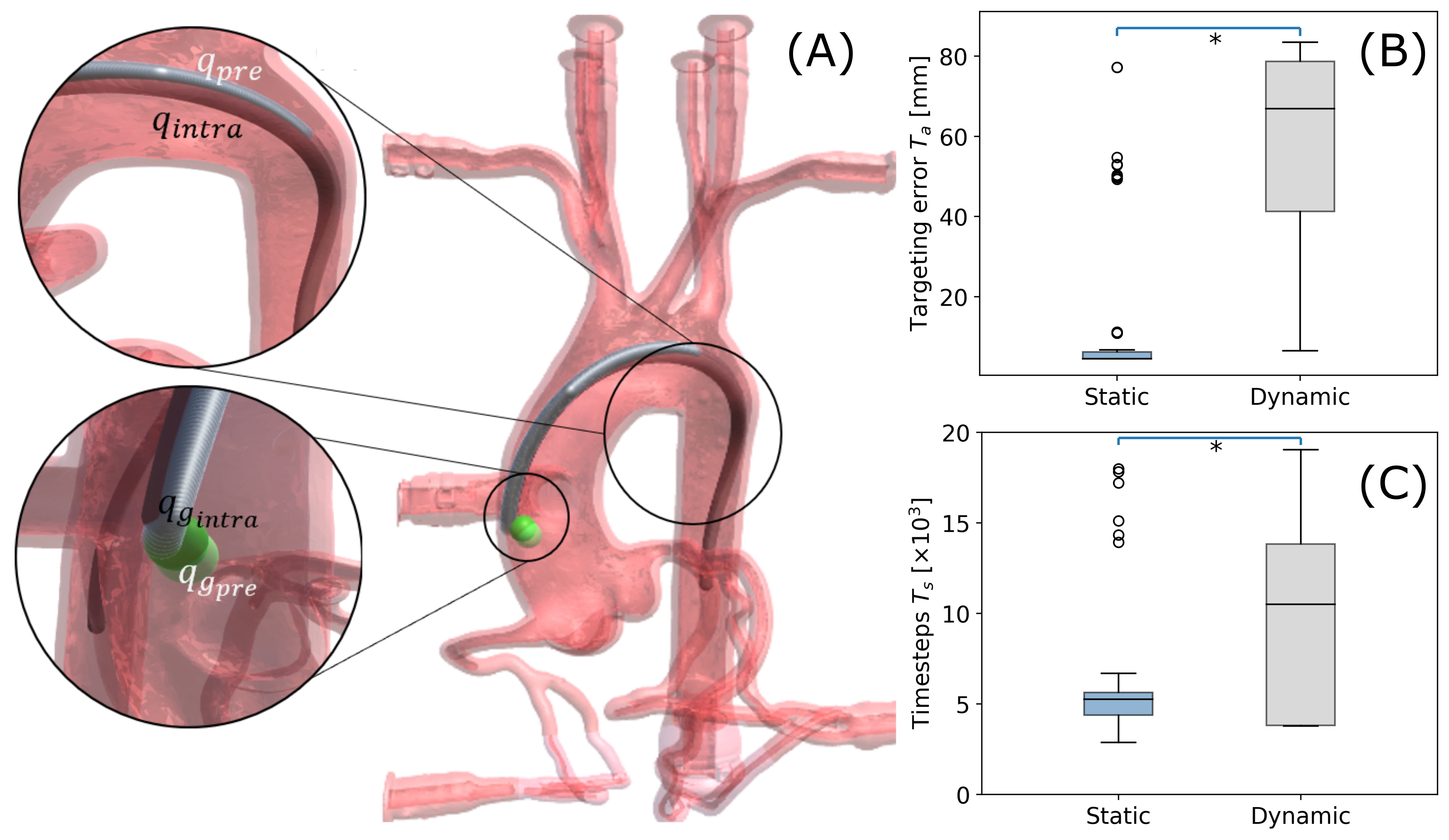}
\caption{\textit{In-silico} performance of the proposed \ac{cgail} network in static and dynamic environments: (A) The intra-operative adaptation of the pre-operative path $q_{pre}$ when the environment is deformable and the target moves; (B) Targeting error $T_a$ and (C) Timesteps $T_s$ in static and dynamic environments. ($*, p<0.05$ using Kruskal-Wallis test)}
\label{fig:comp_softrig}
\end{figure}

\subsection{\textit{In-silico} Path Planning}
The performance comparison of the proposed \ac{cgail} and the state-of-the-art GAIL \cite{chi_2020} in a static environment is presented in Fig.~\ref{fig:comp_path}. The number of experiments for comparison is set to 100, and results are reported regardless of whether the target was successfully reached. Fig.~\ref{fig:comp_path}A shows an example of trajectories obtained by these two approaches. The \ac{cgail} approach generates paths with smaller targeting errors (see Fig.~\ref{fig:comp_path}B). An ablation study of the \ac{cgail} is also presented, investigating its performance when $\mu=0$, $\lambda=0$, or $\nu=0$. If we set the threshold $\epsilon=10$, the success rate of the \ac{cgail} network, originally at 80\%, decreases to 11\%, 58\%, and 48\% for the scenarios when $\mu=0$, $\lambda=0$, and $\nu=0$ respectively. The ablation study results, as shown in Fig.~\ref{fig:comp_path}B, indicate that \ac{bc}, \ac{gail} and curiosity modules are all critical for accurate target reaching. Compared to the \ac{ppo}+\ac{gail} framework utilized in neurosurgery, as detailed by Segato \emph{et al.} \cite{segato2021inverse}, our \ac{cgail} network introduces further advancements by integrating both the \ac{bc} and \ac{cl} modules. The ablation study reveals a significant decline in performance in the absence of the \ac{bc} module, i.e., when $\mu = 0$. Consequently, it is inferred that the method by Segato \emph{et al.} would underperform compared to our proposed approach. Furthermore, the robustness of the proposed network is verified by the success rate of the \ac{cgail} network, that is 80\%, compared to that of the GAIL network \cite{chi_2020} which is only 42\%, when $\epsilon=10$. However, compared to \ac{gail}, \ac{cgail} requires more timesteps to reach the target (see Fig.~\ref{fig:comp_path}C). This indicates that either the path is longer or there are more samples taken along the path to achieve the same level of path length. The duration is affected by a large number of failures that often cause early termination of the experiment. A common failure is that the path planning tends to stop due to non-minor collision with the aortic arch and are therefore unsuccessful attempts. Minor collisions that cause slight deformation within reasonable stress ranges were considered harmless \cite{ye2016fast}. In our prior simulator study \cite{li2022position}, the maximum permissible force applied by the user varies with the anatomical structure involved. For instance, in the aorta, the maximum force is capped at 0.8N. An absolute collision force is determined using Newton’s Second Law of Motion. A collision is classified as minor if the resultant force is below the maximum threshold and there is no penetration.

The agent was then trained using pre-trained weights from the static environment in a dynamic environment, where deformations are caused by contact with the catheter and the heartbeat motion. Fig.~\ref{fig:comp_softrig}A demonstrates the ability of real-time planning in a complex dynamic environment with deformable vessels and a moving target. The top zoom-in view shows that the paths of pre-trained ${\bm{q}}_{pre}$ and intraoperatively trained ${\bm{q}}_{intra}$ are initially close and then separate in the aortic arch, where the degree of deformation is greater. The bottom zoom-in view shows the change from the initial target position ${\bm{q_g}}_{pre}$ to the intraoperative target position ${\bm{q_g}}_{intra}$. Fig.~\ref{fig:comp_softrig}B-C present performance comparisons in static and dynamic environments. The timesteps are much more in the deformable environment. This is due to the fact that there are more possibilities when interacting with a deformable environment than with a static one. Due to the greater complexity of navigation and low success rate, targeting error is larger and more widely distributed than in the static environment. Nevertheless, when considering only the successful trials, the targeting error is observed to be 4.76$\pm$0.57mm in the static environment and 7.57$\pm$0.65mm in the dynamic environment, respectively, demonstrating comparable results. The success rate of 17\% is relatively low but indicates that it is possible to find a feasible path in a constrained environment, compared to the success rate of 0\% for the \ac{gail} network \cite{chi_2020} in the dynamic environment. A common failure is the inability to reach the moving targets with a distance threshold of 10mm. Including this distance threshold to reach a target in the curriculum module would improve network performance, such as success rate. This would be investigated at a later stage. The limited bending capability of robotic catheters and the presence of moving targets in deformable environments led to a reduced success rate within a limited number of timesteps. Despite the relatively low success rate of \ac{cgail}, it has outperformed the state-of-the-art approach.

One limitation of our proposed method, particularly in medical applications, is the exclusion of blood fluid flow dynamics in our developed simulator. To address this, future iterations should incorporate fluid dynamics, potentially using subject-specific geometries as suggested in \cite{schwarz2023beyond, jianu2024autonomous}. The presence of blood introduces friction and resistance against the catheter as it moves through the vessels \cite{wagner2021bio}. Integrating fluid dynamics could yield more precise predictions of vessel deformation and catheter movement. Such enhancements are likely to elevate the realism of the simulated scenarios, thereby improving the accuracy of the path planning to better reflect real-world conditions.

\subsection{\textit{In-vitro} Validation}
To essentially eliminate the impact of the user's learning curve, the user is required to be pre-trained. For example, the learning curve of the targeting error by following the \ac{cgail} path decreased from 3.6 to $\sim$2mm in ten trials. The user can achieve a targeting error of 0.23mm in the later stage. Since the learning curve converges to a plateau, it can be concluded that the user has stable catheter manipulation ability. A video is made available showing example experiments of following the \ac{cgail} path and the centerline \footnote{\href{https://youtu.be/GhBi_xHTMFw}{\color{blue}{\nolinkurl{https://youtu.be/GhBi_xHTMFw}}}}.

\begin{figure}[tb]
\centering
\includegraphics[width=\linewidth]{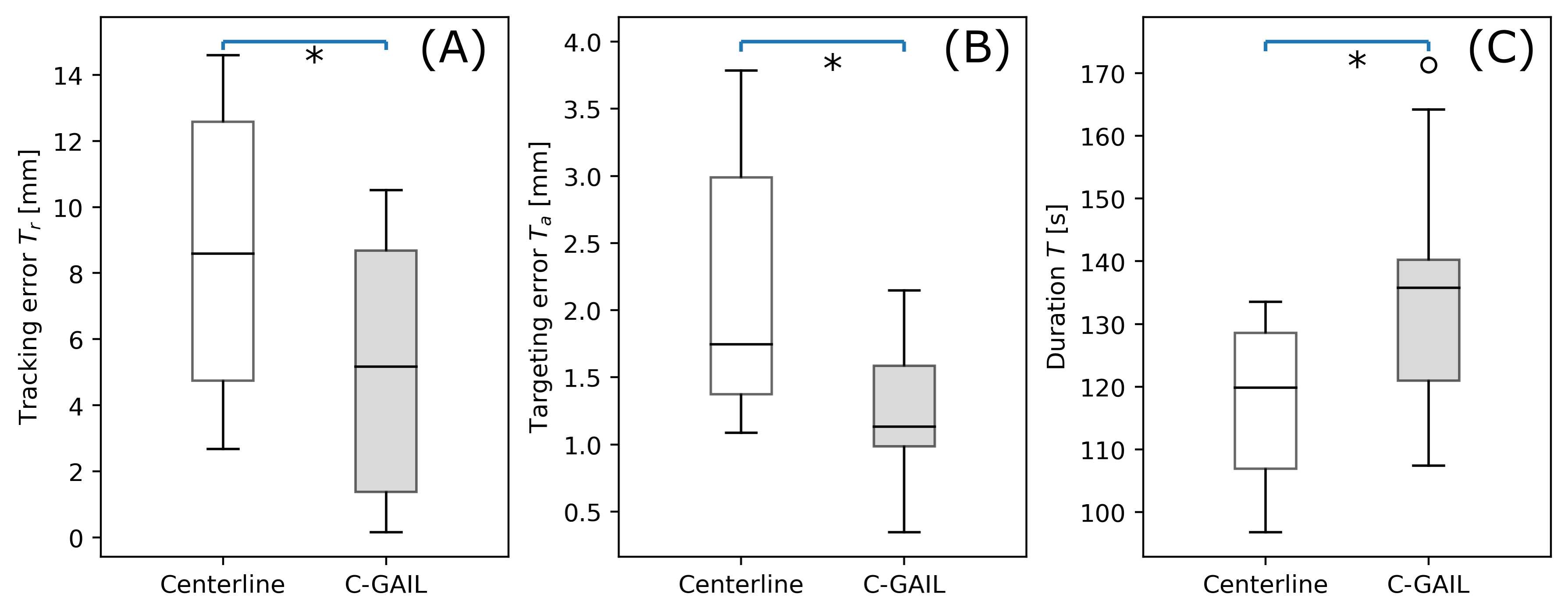}
\vspace{-10pt}\caption{\textit{In-vitro} performance comparison between \ac{cgail} path and centerline, with respect to (A) tracking error $T_r$, (B) targeting error $T_a$, and (C) duration $T$. ($*, p<0.05$ using Kruskal-Wallis test)}
\label{fig:expTest}
\end{figure}
\begin{figure}[tb]
\centering
\includegraphics[width=0.9\linewidth]{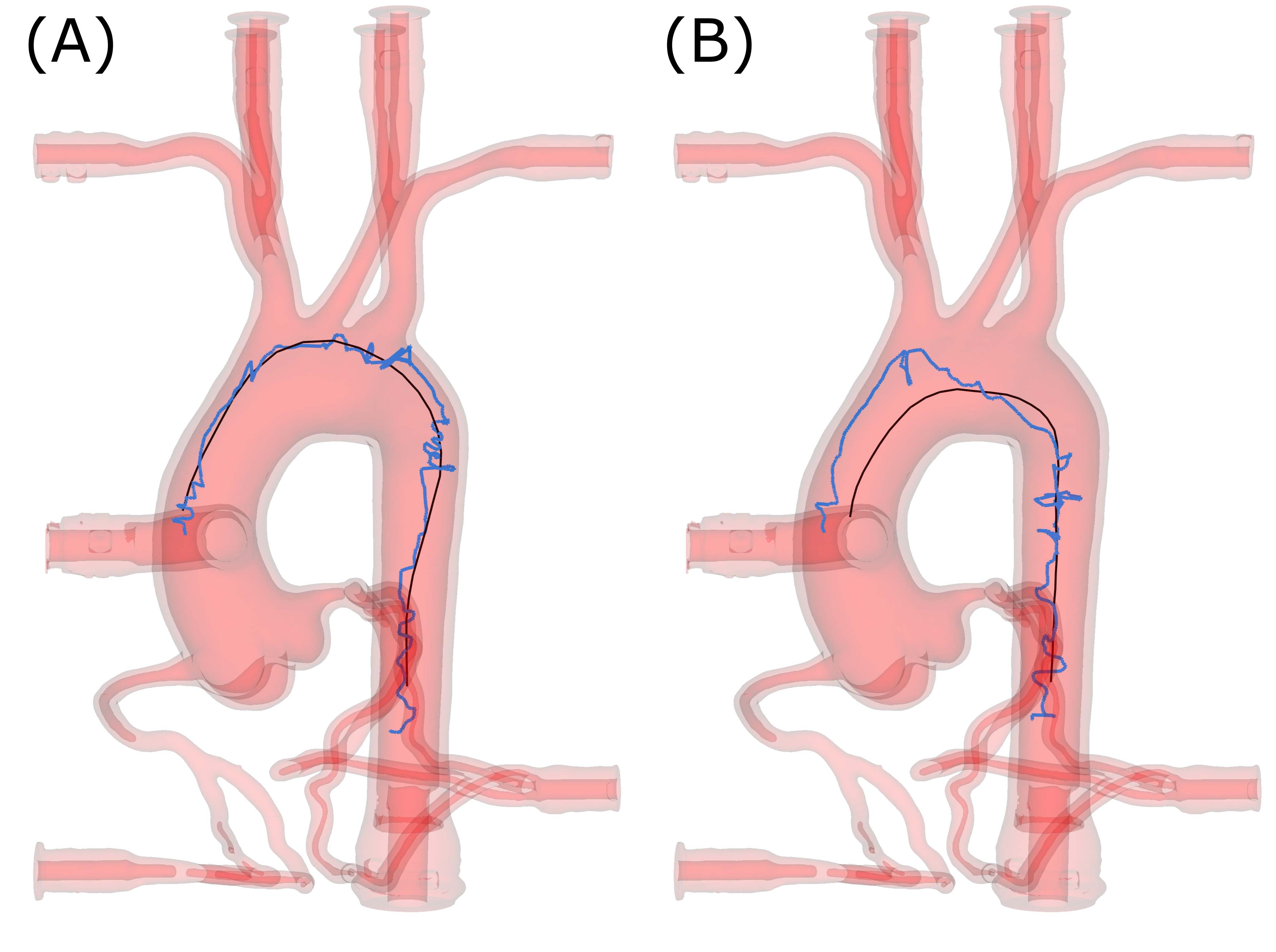}
\caption{Representative \textit{in-vitro} trajectories (blue) and path guidance (black) using: (A) \ac{cgail} path, and (B) centerline.}
\label{fig:expTraj}
\end{figure}

The performance comparison between the \ac{cgail} path and the centerline is shown in Fig.~\ref{fig:expTest}. The \ac{cgail} path leads to a smaller tracking error of 5.18$\pm$3.48mm. It confirms that \ac{cgail} leads to a path that can be followed better by the robotic catheter. Furthermore, the \ac{cgail} path also shows a smaller targeting error of 1.26$\pm$0.55mm (Fig.~\ref{fig:expTest}B). Clinicians typically indicate an accuracy range of 1-3mm as acceptable \cite{nijland2017evaluation, bourier2015sensor}. Therefore, the mean targeting error of 1.26mm effectively meets these clinical accuracy requirements. Following the \ac{cgail} path takes a longer time than following the centerline. This is interpretable given that since the user learned that the centerline waypoints could not be reached anyway when passing the aortic arch, the user kept inserting the catheter at the maximum bending angle and spent less effort following the centerline. While in the other scenario, the user realized that (s)he could follow the \ac{cgail} path waypoints. Therefore, it takes a longer time than following the centerline. Two of the ten experiments with the \ac{cgail} path took longer because the user made a retraction motion in order to re-reach the target more accurately. Both the \ac{cgail} path and centerline following experiments had a 100\% success rate. To reduce tracking and targeting errors, future efforts could explore the use of control input devices offering a broader range of motion than joysticks, such as haptic devices.

Figure~\ref{fig:expTraj} presents a representative graphical comparison of actual trajectories and path guidance in two scenarios: following the \ac{cgail} path and following the centerline. The trajectories of the catheter tip following the \ac{cgail} path are more stable and smoother, with a curvature of $3.1\pm26.6$mm$^{-1}$, compared to a curvature of $3.5\pm38.9$mm$^{-1}$ when following the centerline. In the other scenario, the catheter tip can follow the centerline well at the beginning. However, the trajectory then moves away from the centerline as the catheter passes through the aortic arch due to the limited bending capability. It is important to clarify that the trajectory refers to the movement of the catheter tip. In the \textit{in-vitro} environment, where gravity plays a role, contact between the catheter body and the vessel walls is unavoidable. However, such contact poses minimal risk due to the catheter body's non-sharp nature. The focus of this study does not extend to the catheter's back-end configuration, which may be pertinent for advancing to higher levels of autonomy, such as achieving autonomous navigation along the vessels \cite{pore2023autonomous}. For this objective, capturing the entire catheter's shape using \ac{fbg} could prove beneficial.

In this research, the robotic catheter operation was facilitated through teleoperation. The control technique does not influence the path planning phase because the catheter's constraints, including its bending capabilities, have been preemptively incorporated. Additionally, this dimension was further investigated in a subsequent study \cite{wu2024comparative}, wherein 15 participants with diverse experience levels evaluated the system's effectiveness using C-GAIL path planning within various modes of interaction.

\subsection{Transferability}
\begin{figure}[tb]
\centering
\includegraphics[width=\linewidth]{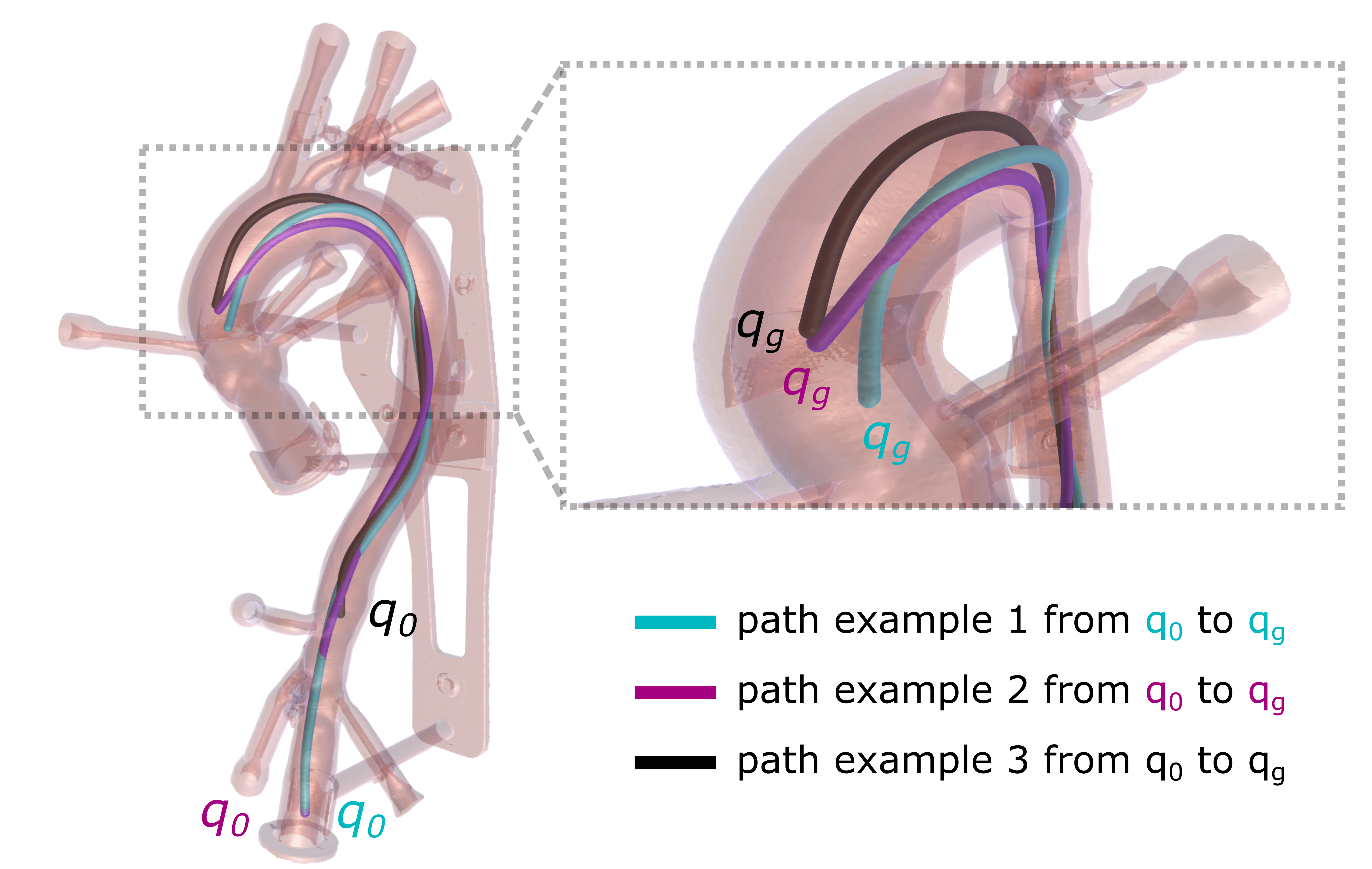}
\caption{Examples of \textit{in-silico} \ac{cgail} paths in a different aortic anatomy  (Materialise NV, Leuven, Belgium).}
\label{fig:long_model}
\end{figure}

To further generalize the proposed path planning framework, \textit{in-silico} experiments were conducted in a different aortic anatomy (Materialise NV, Leuven, Belgium). 70 demonstrations were recorded, and 100 testing trials were performed. The success rate of the proposed \ac{cgail} network was 79\% when $\epsilon=10$. Other performance values with respect to targeting error and timesteps are summarized in Table~\ref{tab:models}. Representative paths obtained by the \ac{cgail} approach are shown in Fig.~\ref{fig:long_model}. These results indicate the robustness and feasibility of applying the proposed path planning framework in anatomies with different geometries.

This paper employs an \ac{em} sensor to measure the tip pose, which has been validated in a phantom study. \ac{em} sensor has demonstrated its efficacy also in \textit{in-vivo} environments \cite{ourak2021fusion, ha2021robust} considering its small size ($<$1mm in diameter and $<$10mm in length) and bio-compatibility. Furthermore, other sensors, such as \ac{fbg} sensor, as well as imaging modalities, can also be utilized to obtain the catheter tip pose and the body shape. In certain specific cases (e.g., free space), the catheter tip pose can also be obtained with a sensorless approach, e.g., an accurate forward kinematic model if control variables are provided \cite{wu2021hysteresis}. One limitation of this study is that there was no intra-operative environment reconstruction for the \textit{in-vitro} experiments. Instead, a simulated deformable model was utilized to predict intra-operative deformations. In clinical settings, real-time vessel reconstruction from \ac{ivus} or \ac{oct} images would be suitable for providing raycast observations elaborated in this work. 

\begin{table}[tb]
\centering
    \caption{In-silico performance of the proposed \ac{cgail} network in a different aortic anatomy.}
    \label{tab:models}
\begin{tabular}{p{.03\linewidth}p{.06\linewidth}p{.03\linewidth}p{.04\linewidth}p{.05\linewidth}p{.03\linewidth}p{.06\linewidth}p{.04\linewidth}p{.04\linewidth}p{.03\linewidth}p{.02\linewidth}} 
\toprule
\multicolumn{5}{c}{$T_a$ (mm)} & \multicolumn{5}{c}{$T_s$ ($\times10^3$)} & \multicolumn{1}{c}{$\delta$} \\
\cmidrule(lr){1-5}\cmidrule(lr){6-10}
25th & median & 75th & mean & std & 25th & median & 75th & mean & std & \multicolumn{1}{c}{(\%)}\\
\hline\noalign{\smallskip}
6.58 & 6.84 & 8.57 & 17.98 & 23.51 & 9.41 & 10.22 & 11.59 & 9.71 & 2.78 & 79\\
\bottomrule
\end{tabular}
\end{table}

\section{Conclusion}
This study set out to design a robust path planning approach respecting the kinematics of robotic catheters and real-time changes in deformable cluttered environments. This path planning approach uniquely accounts for both the deformable nature of the environment and the dynamic movements of the target, distinguishing it from existing methods. \textit{In-vitro} experiments and an extensive follow-up user study \cite{wu2024comparative} underscore the algorithm's feasibility in generating suitable paths that align with the actual steering capability of the catheter and the deformable environment, thereby enhancing navigation support with greater accuracy.

The insights gained from this study add to the rapidly expanding field of autonomous navigation for robotic catheters. The findings of this study suggest that the proposed path planner can effectively handle the uncertainty present in vessel deformation.

A limitation of our study is the absence of intra-operative environment reconstruction during the \textit{in-vitro} experiments. In clinical practice, real-time reconstruction of vessels using \ac{ivus} or \ac{oct} images would be advantageous for generating the raycast observations discussed herein. Further clinical studies that involve real-time vessel reconstruction will be carried out. Additionally, future efforts could benefit from integrating blood fluid flow dynamics into the simulator, thereby increasing the realism of simulated scenarios and enhancing path planning accuracy.

\bibliographystyle{IEEEtran}
{\footnotesize \bibliography{main}}

\end{document}